\documentclass[sigconf]{aamas} 

\usepackage{balance} 

\usepackage{subcaption}     
\usepackage{ifsym}          
\usepackage{algorithm}      
\usepackage{algpseudocode}  
\algdef{SE}[DOWHILE]{Do}{doWhile}{\algorithmicdo}[1]{\algorithmicwhile\ #1}

\acmSubmissionID{1129}

\title[Human Values in Multi-Agent Systems]{Human Values in Multi-Agent Systems}

\author{Nardine Osman}
\affiliation{
  \institution{Artificial Intelligence Research Institute (IIIA-CSIC)}
  \city{Barcelona, Catalonia}
  \country{Spain}
  }
\email{nardine@iiia.csic.es}

\author{Mark d'Inverno}
\affiliation{
  \institution{Goldsmiths, University of London}
  \city{London}
  \country{United Kingdom}
  }
\email{dinverno@gold.ac.uk}

\begin{abstract}
One of the major challenges we face with ethical AI today is developing computational systems whose reasoning and behaviour are provably aligned with human values. 
Human values, however, are notorious for being ambiguous, contradictory and ever-changing. 
In order to bridge this gap, and get us closer to the situation where we can formally reason about implementing values into AI, this paper presents a formal representation of values, grounded in the social sciences. 
We use this formal representation to articulate the key challenges for achieving value-aligned behaviour in multiagent systems (MAS) and a research roadmap for addressing them.
\end{abstract}


\newcommand{\BibTeX}{\rm B\kern-.05em{\sc i\kern-.025em b}\kern-.08em\TeX}

\begin{document}


\pagestyle{fancy}
\fancyhead{}


\maketitle 


\section{Introduction}

It is widely recognised that computational models of human values are critical for designing ethical multi-agent systems (MAS) involving mixed communities of humans and artificial agents~\cite{russell2019human,GFAIH21,10.1007/s11023-020-09539-2,Weide2010}. 
We propose an intuitive, foundational and concrete representation of values, grounded in the social sciences, required to build the primitive computational mechanisms needed for reasoning about values in MAS. 
We believe that no such model exists in the published literature. 
With our proposed formal representation, we show how we can set out the computational challenges of building MAS with value-aligned behaviours. 
Through our efforts to draw on work from other disciplines and the social sciences, in particular, we have intentionally set out to pave the way for interdisciplinary research teams to come together under a shared conceptual underpinning. 
To support this, we show how our formal model can be used to define four key research challenges for building ethical MAS, which are set out as follows:
%


\begin{enumerate}
    \item Challenge 1. How can we identify the values we are dealing with? -- the value identification and categorisation problem.
    \item Challenge 2. How can we move from individual to collective values? -- the value aggregation and agreement problem.
    \item Challenge 3. How can agents decide what they value now and what they do next? --  the value-aware decision-making problem. 
    \item Challenge 4. How can we build sustainable value-aligned multi-agents systems? -- the ethical MAS problem.
\end{enumerate}


\section{A Formal Model for Value Representation}\label{sec:valueTax}

Our stance on what values are is aligned with the social sciences where they are abstract concepts that guide behaviour, but whose exact meaning and interpretation varies heavily with context and/or time~\cite{Rohan2000,Schwartz2012AnOO}. 
(The modelling approaches used in the social sciences vary but we have set out to develop a model which draws from this range as much as possible.)
%
However, in order to achieve any meaningful evaluation of values, a concrete computational-representational understanding of values is required. 
That is to say that whilst we might talk about fairness as a value we want to have in general, in a specific community, fairness would need to be defined more concretely. 
For example, in a system we have recently implemented to support mutual aid communities, fairness is understood to be: ``any member does not ask for significantly more help than the help they have volunteered for others''. 
%
%
While the former is an abstract concept, the latter is a concrete shared understanding (meaning) attached to the value fairness through a property whose satisfaction (or degree of satisfaction) can be automatically verified. 
This idea of moving between an abstract value to a specific rule-based implementation leads us to propose that values be defined using taxonomies. 
Any general value-concept (such as fairness) then becomes more specific as we move down the taxonomy, and becomes concrete, computational and verifiable at leaf nodes. (This approach is consistent with the work in value-sensitive design~\cite{vandePoel2018} on value change taxonomies.)

Another important concept considered to be core in the social sciences is that of \emph{value importance}, where the relative importance of an individual or community's values is what guides behaviour. 
We incorporate this concept by attaching a measure of \emph{importance} to each node of the value-taxonomy, without stating what form that measure might take. 

Our formal proposal for value representation through a taxonomy is given in Definition~\ref{def:valueTaxonomy}. 
\begin{definition}[Value-taxonomy]\label{def:valueTaxonomy}
A value-taxonomy $\mathcal{V}=(N,E,I)$ is defined as a directed acyclic graph, where: 
\begin{enumerate}
    \item The set of nodes $N=N_{l}\cup N_{\phi}$ represents \emph{value-concepts}, and is composed of two types of nodes: i) those that are specified through labels, with $N_{l} \subset \mathrm{L}$ representing the set of \emph{label-nodes} and $\mathrm{L}$ is the set of all \emph{value-labels} representing abstract value-concepts like `fairness' or `reciprocity'; and ii) those that are specified through concrete properties, with $N_{\phi} \subset \Phi$ representing the set of \emph{property-nodes} and $\Phi$ representing the set of all \emph{value-properties} whose satisfaction can
    be automatically verified at different world states, such as having the number of times one asks for help in a mutual aid community to be no bigger than 125 \% of the number of times one has given help. 
    \item The set of edges $E : N \times N $ is a set of directed edges $(n_{p},n_{c}) \in E$ that represent the relation between value-concepts $n_{p}$ and $n_{c}$ (the parent and child nodes, respectively) illustrating that the value-concept $n_{p}$ is a more general concept than $n_{c}$. 
    \item The importance function $I: N \to CD$ assigns -- for each value-concept in $N$ -- an importance value from the codomain (range) $CD$.
\end{enumerate}
\end{definition} 
%

%
%

We argue that the property-nodes of the value-taxonomy allow for a computational approach to reasoning about values, and to the problem of value alignment (the higher the satisfaction of a value's properties, the higher the alignment with that value -- more on this in Section~\ref{sec:roadmapVdesign}). 
The importance of nodes allows for value-aware decision-making that includes which actions and which norms to abide by (more on this in Sections \ref{sec:roadmapVdecisions} and~\ref{sec:roadmapVdesign}). 
Finally, the structure of the taxonomy allows for different interpretations of values in different contexts (more on this in Section~\ref{sec:valueIdentification}). It also allows for reasoning and deliberation about the meaning of values (more on this in Sections \ref{sec:valueIdentification} and~\ref{sec:roadmapVagreement}).

Our ongoing work [{\bf blinded reference}] makes a case for our proposal for representing values, motivating both the need and academic significance for such a proposal and details the alignment of our formal proposal with key research from the social sciences. 
The work also provides instances of implementations that could be chosen by any designer alongside mechanisms and algorithms that 1) ensure coherence of value-importance in a value-taxonomy and 2) allow for the implementation of computational value alignment models. 
This blue sky paper, on the other hand, uses our proposed value representation to set out the key research challenges for achieving computational value alignment in MAS, and through this proposes a roadmap for future interdisciplinary research. 


%
%
%
%

\section{Roadmap for achieving value aligned behaviour in MAS}\label{sec:roadmap}
\subsection{The value identification \& categorisation problem}\label{sec:valueIdentification}

Value identification and categorisation is the challenge of establishing what the values are in any MAS, and to identify their inter-relationship and their importance for any current or imagined multi-agent system. 
There are two parts to this. First, if we wish to join an organisation and understand how to be successful within it, we will need to understand the values by which that system operates and how they relate to our own. Similarly, if our challenge is to design a new multi-agent system, then we will want to work with all stakeholders to identify the values to be upheld within the operation of that system. 


\paragraph{Related work.} Current work in AI on this topic aims at eliciting and learning relevant values from (typically, written records of) people's interactions. Natural language processing techniques are being used to estimate, in a (semi-) automatic manner, underlying human values from text. 
For instance, \cite{Liu2019PersonalityOV} provides an analysis of values based on words used in e-commerce reviews, and \cite{Lin2018AcquiringBK} estimates relevant values in tweets by combining textual features and context knowledge from Wikipedia. 
However, these techniques are employed only once a predefined high-level value list has been selected, such as the well-known Schwartz value system~\cite{Schwartz2012AnOO}. 
Using any pre-defined fixed list is a limitation not only in the assumption that the list is appropriate for the context, but it also prevents values from changing over time, a view we share with the value-sensitive design community~\cite{vandePoel2018}. 
Amongst the approaches that do not start with a predefined value list but sets out to identify the relevant values can be found in~\cite{89b26068890444aa88b4a15afe36625c}, which presents a crowd-powered algorithm to generate a hierarchy of general values. 
Another such  can be found in Axies, using human and automatic techniques for identifying context-specific values using natural language processing~\cite{10.5555/3463952.3464048}. 

\paragraph{Identifying the way forward in a roadmap for future research.} 
The understanding of values in these existing approaches typically remains at an abstract level. 
They are articulated through textual headings (such as `fairness'), without further exploring the concrete meaning of each of these listed values, and no mechanism for deliberating and reasoning about these value lists.  
We will provide high-level descriptions of these overlooked mechanisms as a foundation for further research and development. 
These descriptions then motivate questions on the meaning of values (through property nodes), and the relations between different value labels. 
Specifically, we identify some of the key research challenges ahead:

\begin{enumerate}
    \item Extending existing research on value identification (e.g.~\cite{10.5555/3463952.3464048}) 
    so that relations between those values initially identified by human/AI processes can be established, resulting in constructing a value-taxonomy as we propose.
    \item Developing mechanisms for constructing property-nodes for values, usually context dependent, and link those property nodes to the abstract label-nodes. This is crucial for any computational approach to building AI systems that can explicitly reason about values. 
    \item Developing automatic propagation mechanisms that, given the importance of some set of nodes within a taxonomy, can calculate the importance of all the remaining nodes, and doing so in such a way which ensures coherence of importance across the whole taxonomy. 
    Developing such mechanisms will be useful in practice because obtaining the importance of every single node is usually not straightforward (see discussion below).  
\end{enumerate}

We believe that addressing these challenges is necessary for real progress in the practice of introducing values into AI systems. 
Until we can make progress with these research challenges, it is difficult to see how we will trust AI systems to be able to truly operate according to our values -- the critical ethical concern of AI. 
Identified value taxonomies provide an explicit mechanism for reflecting human values of relevant stakeholders, where these taxonomies can be seen and checked by those stakeholders. 
It is not straightforward for humans to explicitly specify their value taxonomies. While many ethicists working in the field of value-sensitive design have been explicitly eliciting the important values and their inter-relationships from stakeholders, asking that the users of technologies undertake such a process would be too demanding and time-consuming in practice. 
We can expect the typical user/stakeholder to have a broad understanding of what an AI system has learned of their value systems and the explicit way it has chosen to model them (i.e. the constructed value taxonomies). Moreover, we can expect them to approve or disapprove various aspects of the learned value systems, and so, guide AI in the way it learns and represents values. 
What we cannot expect is for the layperson to get into the details of the value importance of each node, the exact relationships between nodes, etc. 
So the balance can only be addressed through the collaboration of AI and human stakeholders, where the AI informs the human of what it is learning, and the human's input can help guide the learning process. 


\subsection{The value aggregation \& agreement problem}\label{sec:roadmapVagreement}
While value identification and categorisation focuses on identifying the important values of a single entity (e.g. human, community, organisation, company, etc.), value aggregation and agreement focuses on the mechanisms required for constructing the value-taxonomy of a collective. The question is how do we move from a set of individual value taxonomies to collective ones? 

\paragraph{Related work.} 
\cite{10.1007/s11023-020-09539-2} argues that we live in a pluralistic world with different entities holding different value systems. To ensure behaviour in a MAS is aligned with human values, decisions are needed about the value system of any MAS. 
To arrive at that value-system potential conflicting value systems of individuals or even sub-groups of individuals needs to be addressed. \cite{10.1007/s11023-020-09539-2} defines this problem as identifying the value system that receives ``refective endorsement despite widespread variation in people's moral beliefs''. \cite{Pigmans2017,Pigmans2019} highlight the challenges of addressing conflicting individual interests in the field of water policy-making and reports how deliberation around the value systems of different stakeholders can help address such conflicts. 
Some work in this field~\cite{Lera-LeriBSLR22} makes use of computational social choice to aggregate individual value systems and yield a consensus value system. This approach considers a range of ethical approaches, from utilitarian (maximum utility) to egalitarian (maximum fairness). 

\paragraph{Identifying the way forward through a roadmap for future research.} Whilst research on value aggregation and agreements is beginning to emerge, many challenges still need to be addressed, including the following, which arise more clearly now we have provided a formal, concrete model for value-systems:

\begin{enumerate}
    \item Developing mechanisms for computational social choice. These can take into consideration the meaning of values as defined using the property nodes of our formal proposal for value systems. 
    In other words, a complex aggregation mechanism is needed, not to aggregate the value importance of individual value-concepts, but to aggregate entire value taxonomies into an aggregated value-taxonomy.
    
    \item Developing mechanisms for value agreements. In addition to aggregation mechanisms that compute the value system of a collective, agreement technologies (such as argumentation and negotiation mechanisms) are required to support the constituent individual's reaching an agreement on the adopted value system of a proposed collective by deliberating over the meaning (property-nodes) and importance of values. 
   %
\end{enumerate}

We note that in both these challenges the individual's value system may or may not change, since the focus is on agreeing on a value system for the collective. 
As such, conflicts between individual value systems and the system of the collective might arise. 
If the degree of incoherence is sufficiently strong, this may trigger the individual to take no further part in that collective and look for alternatives better aligned with their own value system. 
In other situations, an individual agent might be obliged to interact within the collective, and so recognise the value-system of the community (regardless of whether they decide to take actions that adhere or not to the value system of the collective). 

\subsection{The value aware decision making problem}\label{sec:roadmapVdecisions} 

Identifying the value systems of individuals and collectives as discussed, provides the basis for reasoning over values. 
Armed with the knowledge of its own value system and that of the collective in which it is currently acting, the agent can reason about how to behave. 
The computational challenge is concerned with developing enhanced decision-making mechanisms that take different value systems, especially the individual's and the collective's, into consideration. 

\paragraph{Related work.} 
In the field of value-driven decision-making, persuasion has been one approach to motivate an agent to act in a specific way. 
In \cite{Bench-Capon2009}, an argumentation framework is presented where the stance is that persuasion relies on the strength of arguments, which depends on the social values which are advanced. 
In \cite{TostoD12}, an agent model is described where agent actions are driven by both their needs and their values, where values are used to prioritise those needs. 

In other work~\cite{ChhogyalNGD19}, the notion of trust has been explored as a mechanism for influencing decision-making, where the past reliability of an agent's actions is used to decide whether that agent can be trusted or not. 
The argument is made that when past experiences cannot be used to assess the reliability of others, the sharing of values between the trustor and trustee can help, and an approach is developed to evaluate trust based on the degree to which shared values can be established. 
In~\cite{ijcai2017p26}, reasoning about values is used to help agents make choices over plans to adopt.

\paragraph{Identifying the way forward through a roadmap for future research.} As illustrated earlier, our stance is taken from the social sciences where values are abstract concepts that guide behaviour. 
%
%
Some of the challenges in this area can now be identified more clearly:  
\begin{enumerate}
    \item Investigating reasoning about actions that include specific recognition of the importance and (of course) the relative importance of those values relevant to behavioural choice. 
    Different approaches could be investigated here, such as adopting practical reasoning in cognitive agent models or extending existing BDI models to include value-taxonomies. 
    One may also investigate a value-enhanced theory of mind, where agents can observe each others' actions, build a model of each other accordingly using theory of mind, and reason about those actions and their underlying intentions. This process is undertaken to support the observer's own decision-making processes. 
    The main focus of these mechanisms will be on incorporating value-taxonomies in order to reason about the underlying values driving others' behaviour and make value-aware decisions accordingly.  
    While, up until now, values have been mostly used as labels in the literature without a real understanding of a value's meaning, using value-taxonomies can enrich such reasoning mechanisms.
    \item Developing value-driven deliberation mechanisms that influence behaviour through persuasion, argumentation, or negotiation. This would extend existing work, such as that of~\cite{Bench-Capon2009}, with more work on value agreement. 
    For example, instead of persuading how one should act based on the value alignment of those actions, one might try to persuade or argue about the value system itself and how it could be updated. 
    Convincing others to change their value taxonomies can be used to persuade that individual to act in a certain way. 
    %
    Research will focus on deliberation about the nodes in a value-taxonomy making reference to their importances.
    \item Developing greater explainability mechanisms to help humans understand and investigate the \emph{ethical} implications of their own actions in terms of the impact they might have, as well as to better understand the ethical motivations driving agents to act in specific ways. 
    %
    This requires mechanisms for reasoning about the possible implications of chosen actions alongside an understanding of the concrete meaning of values, as provided by our value taxonomies and their property leaf nodes.
    %
    %
\end{enumerate}


\subsection{The ethical multiagent system problem}\label{sec:roadmapVdesign}

While the third challenge is focused on the value alignment of an individual's decision-making process, this fourth challenge concerns 
%
developing MAS in such a way that their alignment with human values can be evidenced as holding over a sustained period of time, especially as different value systems from users and stakeholders may evolve.

\paragraph{Related work.} 
The design of technologies that are aligned with our human values is a well-established field in the social sciences, known as value-sensitive design (VSD)~\cite{10.1561/1100000015}. 
Whilst VSD relies on offline participatory design alongside offline evaluations, our programme of research complements this approach by providing an online verification mechanism that computationally assesses the degree of alignment. 
 %


Since norms have been traditionally used in MAS to mediate behaviour, proposed mechanisms that assess a MAS's alignment have been reduced to assessing the value alignment of the MAS's norms. If a set of norms bring about outcomes that are more aligned with a given value system, the set of norms and its corresponding MAS are said to be aligned with that value system. The research in this field has mostly focused on choosing an optimal set of norms that optimise the value-alignment of the MAS~\cite{SerramiaLR20,MontesS21}. 
%
%
In~\cite{SerramiaLR20}, norm synthesis is automated, and it is based on some preliminary knowledge of which norms promote which values. 
The work in~\cite{MontesS21} proposes a value-promoting norm synthesis approach that essentially optimises the value-alignment mechanism proposed in~\cite{abs-2110-09240}. 
In~\cite{abs-2110-09240}, value preferences are understood as preferences over world states and value alignment of a set of given norms is based on the degree to which those norms move us towards preferred states. 

\paragraph{Identifying the way forward through a roadmap for future research.} 
While several mechanisms are being proposed in the field of ethical multiagent systems, many challenges in this area remain to be resolved, such as:  
\begin{enumerate}
    \item Enhancing value alignment mechanisms by considering the computational meaning of values as provided by our value-system taxonomy. 
    Existing value alignment mechanisms suffer from two major pitfalls. The first is that they require a lot of manual work from the human side to specify the meaning of values (the property nodes in our taxonomy) and their importance. 
    We believe this can be addressed by the value identification and categorisation mechanisms described earlier in Section~\ref{sec:valueIdentification}. 
    The second is that many of the existing mechanisms reason over values without an understanding of the meaning of those values. 
    Again we believe that providing a computational meaning of values (through property nodes) enhances the ability to reason about value alignment, resulting in better explanations, which we describe next.
    \item Developing explanation mechanisms that help humans understand why one set of norms is preferred over another with respect to a given value-system. 
    Explanations will be more detailed given the introduction of the computational meaning of values that we have provided. 
    Such explanations could strongly support the design of new MAS systems, as well as policy-making and protocol design in general, such as the design of medical protocols, emergency protocols, or policies for regulating irrigation practices. 
    %
    %
    \item Providing new mechanisms for self-governance in MAS through value-driven norm agreement mechanisms. 
    One of the main challenges of self-governing MAS is reaching agreements on the norms that govern those societies. The objective is to develop mechanisms that support groups of agents to find the best set of norms to mediate their interactions. 
    First, formal analysis of norms, multiagent simulation, and AI-based optimisation techniques are some of the techniques that can be used to explore the space of normative systems, searching for the optimally aligned set of norms. 
    Value-driven deliberation mechanisms can then be developed to support the process of reaching collective agreements on the chosen norms. 
    Explanations, as presented in the above challenge, can 
    further enhance the deliberation mechanisms. 
\end{enumerate}




\balance


\section{Conclusions}
We have presented a formal and foundational representation of value systems that allows for computational reasoning about values in multi-agent systems (MAS). 
%
%
We have presented what we believe to be the most intuitive, high-level model influenced by a range of work from the social sciences. 
However, there will always be other modelling approaches, but by outlining this attempt we would hope to see counter-proposals. 
%

We have then used this new representation to set out the research challenges for achieving value-aligned behaviour in MAS alongside a roadmap needed to address these challenges. 

The dream of ethical design of AI requires a coherent and sustained interdisciplinary research effort. 
We have deliberately set out to align our work with the extensive research on values from the social sciences, not only to ground our formal proposals but to provide a conceptual framework which provides a starting point for where interdisciplinary research can take place.




\begin{acks}
This work has been supported by the EU funded VALAWAI (\#~101070930) and WeNet (\#~823783) projects, and the Spanish funded VAE (\#~TED2021-131295B-C31) and Rhymas (\#~PID2020-113594RB-100) projects.
\end{acks}



\bibliographystyle{ACM-Reference-Format} 
\bibliography{references}

\end{document}